\documentclass{article}

\usepackage[preprint]{corl_2026} 

\usepackage{amsmath,amssymb,amsfonts}
\usepackage{algorithmic}
\usepackage{algorithm}
\usepackage{graphicx}
\usepackage{textcomp}
\usepackage{booktabs}
\usepackage{multirow}
\usepackage{xcolor}
\usepackage{subcaption}
\usepackage{enumitem}

\newcommand{\ie}{\textit{i.e.}}
\newcommand{\eg}{\textit{e.g.}}

\title{DiscreteRTC: Discrete Diffusion Policies \\ are Natural Asynchronous Executors}

\author{
  Pengcheng Wang$^1$, Kaiwen Hong$^2$, Chensheng Peng$^1$,\\ {\bf Katherine Driggs-Campbell}$^2$, {\bf Masayoshi Tomizuka}$^1$, {\bf Chenfeng Xu}$^3$, {\bf Chen Tang}$^4$\\
  UC Berkeley$^1$, UIUC$^2$, UT Austin$^3$, UCLA$^4$\\
  \texttt{wangpc@berkeley.edu, cxu@utexas.edu, ctangac@ucla.edu}
}

\begin{document}
\maketitle
\vspace{-2em}

\begin{abstract}
Unlike chatbots, physical AI must act while the world keeps evolving.
Therefore, the inter-chunk pause of synchronous executors is fatal for dynamic tasks regardless of how fast the inference is.
Asynchronous execution---\emph{thinking while acting}---is therefore a structural requirement, and real-time chunking (RTC) makes it viable by recasting chunk transitions as inpainting: freezing committed actions and consistently generating the remainder.
However, RTC with a flow-matching policy is structurally suboptimal: its inpainting comes from inference-time corrections rather than the base policy, yielding little pre-training benefit, specific fine-tuning, heuristic guidance, and extra computation that inflates the latency.
In this work, we observe that discrete diffusion policies, which generate actions by iteratively unmasking, are \emph{natural asynchronous executors that resolve all limitations at once}: they are fine-tuning free since inpainting is their native operation, while early stopping further provides adaptive guidance and reduces inference cost.
We propose \textbf{DiscreteRTC}, which replaces external corrections with native unmasking, and show on dynamic simulated benchmarks and real-world dynamic manipulation tasks that it achieves higher success rates than ContinuousRTC and other baselines.
In summary, DiscreteRTC is simpler to implement with 0 lines of additional code to enable async inpainting, faster at inference with only $\sim 0.7\times$ computation compared with generating actions from scratch, and better at execution with 65\% higher success rate in real-world hockey defend task compared with flow-matching RTC, and 30\% higher compared with training-time flow-matching RTC.
More visualizations are on our \href{https://outsider86.github.io/DiscreteRTCSite/}{website}.
\end{abstract}

\begin{figure}[!ht]
    \centering
    \includegraphics[width=0.85\linewidth]{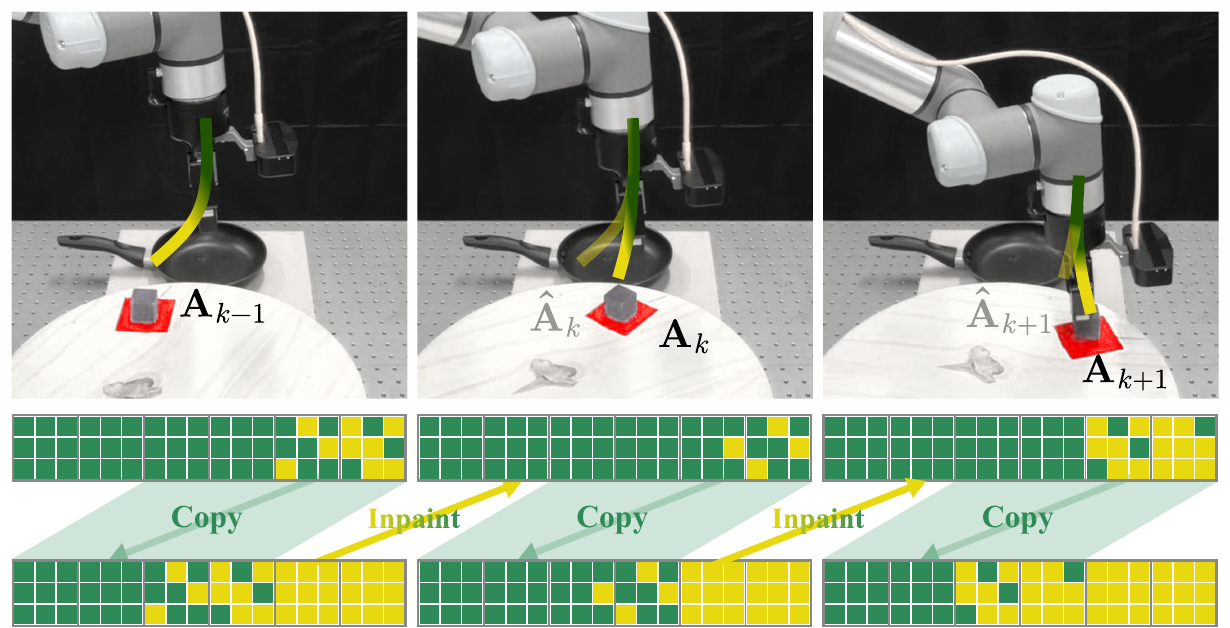}
    \caption{\textbf{Async Execution with discrete diffusion policies solving dynamic manipulation.}
    Gray rectangles and blocks represent the action chunks and the actions.
    Yellow and green cubes represent the masked and unmasked action tokens.
    During each inference cycle, discrete diffusion policies copy the tail of the last action chunk as the committed prefix, and inpaint upon it by simply forwarding itself.
    Compared with flow-matching-based inpainting that relies on $\Pi$GDM, discrete-diffusion-based inpainting inference is simpler to implement, faster at inference, and better at execution.
    }
    \label{fig:cover}
    \vskip -0.2in
\end{figure}


\newpage
\section{Introduction}
\label{sec:intro}
\vspace{-0.5em}

Unlike chatbots~\citep{brown2020language, yang2025qwen3} or image generators~\citep{wan2025wan}, physical AI systems~\citep{black2024pi_0, intelligence2025pi_, bjorck2025gr00t, wang2026dadp, openai2018learning, an2025dexterous, su2025hitter, yu2025skillmimic, hu2026simple, huang2026tic} act with the world evolving: the robot cannot afford to pause while thinking.
This reveals a structural limitation of synchronous execution with action-chunking policies~\citep{chi2025diffusion, lai2025action}---every inference call forces a mandatory pause between chunks during which the robot is idle but the world is not, which is fatal for dynamic tasks such as tracking moving targets~\citep{xie2026dynamicvla}.
Therefore, asynchronous execution---\emph{thinking while acting} over action chunks---is not a latency optimization but a structural requirement.

However, naive async strategies, whether switching chunks immediately or smoothing them via temporal ensembling~\citep{zhao2023learning}, introduce inter-chunk discontinuity or invalid actions because each new chunk is generated in isolation from the committed actions.
Real-time chunking (RTC)~\citep{black2025real} resolves this by recasting chunk transitions as an \emph{inpainting} problem: freezing committed actions from the previous chunk and consistently generating the remainder, representing a simple yet effective approach that has achieved great success in precise and smooth real-world manipulation~\citep{intelligence2025pi_06, intelligence2026pi}.

Specifically, RTC is specially designed for flow-matching policies~\citep{lipman2022flow}.
As the predominant action head architecture in today's state-of-the-art VLAs~\citep{black2024pi_0, intelligence2025pi_, bjorck2025gr00t}, flow-matching has driven much of the recent progress in large-scale robot learning.
However, in this paper, we first systematically show that \emph{\textbf{RTC with a flow-matching head is far from ideal with four critical limitations}}:

\begin{itemize}[nosep]
    \item [(a)] \textbf{Pre-training w/o Inpainting.} Flow-matching policies are not pre-trained with inconsistent noise for inpainting.
    Therefore, scaling pre-training does not directly improve asynchronous performance, and inference-time corrections~\citep{song2023pseudoinverse} are inevitable for inpainting;
    \item [(b)] \textbf{Fine-tuning Required.} As a consequence, adequate inpainting quality demands a dedicated fine-tuning stage with techniques such as action-suffix conditioning~\citep{black2025training, wang2026real} to explicitly introduce the inpainting-specific noise pattern into training;
    \item [(c)] \textbf{Heuristic Guidance.} Moreover, to better leverage the previous action chunks, $\Pi$GDM requires a heuristic schedule fixed across different inference-time cases;
    \item [(d)] \textbf{Extra Inference Cost.} Finally, the correction guidance term at every denoising step roughly doubles inference cost, ironically increasing the very latency RTC aims to hide.
\end{itemize}

Many ongoing efforts~\citep{liu2026learning, lu2026faster, black2025training, yang2026abpolicy} seek to resolve these aforementioned issues within the flow-matching paradigm.
Instead, our key insight and observation are that by replacing the action head with a discrete diffusion policy~\citep{liang2025discrete}, all the aforementioned limitations can be resolved at once. Or, to put it simply: \emph{\textbf{Discrete Diffusion Policies are Natural Asynchronous Executors}}:

\begin{itemize}[nosep]
    \item [(a)] \textbf{Inpainting as Pre-training.} Discrete diffusion policies are pre-trained to inpaint upon randomly masked sequences. Therefore, scaling pre-training directly improves asynchronous performance, and the native forward pass suits inference-time inpainting;
    \item [(b)] \textbf{Fine-tuning Free.} As a consequence, inpainting-specific patterns are implicitly introduced during pre-training, making discrete diffusion a fine-tuning-free approach for high-quality, out-of-the-box asynchronous execution;
    \item [(c)] \textbf{Natural Guidance.} Moreover, with discrete diffusion policies, we can early-exit inference once the necessary action tokens are unmasked, leaving the remaining masking pattern as an adaptive and natural guidance for the next inference;
    \item [(d)] \textbf{Lower Inference Cost.} Finally, with committed tokens from previous chunks, the tokens to unmask per inference are reduced, leading to lower inference cost for inpainting.
\end{itemize}

We propose the resulting async execution method, \textbf{DiscreteRTC}, which replaces the external corrections in flow-matching-based RTC with the native unmasking operation of discrete diffusion policies.
Experiments on the dynamic simulated benchmark Kinetix~\citep{matthews2024kinetix} and real-world dynamic manipulations demonstrate that DiscreteRTC achieves higher success rates than flow-matching-based RTC and training-time baselines.
In summary, DiscreteRTC is simpler to implement with 0 lines of code to enable
async inpainting, faster at inference with only $\sim 0.7\times$ computation compared with generating actions
from scratch while flow-matching-based RTC requires $\sim 1.7\times$ computation, and better at execution
with 65\% higher success rate in real-world hockey defend task compared with flow-matching RTC, and 30\% higher success rate compared with training-time flow-matching RTC.



\section{Preliminaries}
\label{sec:background}


\textbf{Action Chunking Flow Policy.}
We consider an action chunking policy $\pi$ that receives an observation $\mathbf{o}_t$ and outputs a chunk of $H$ future actions $\mathbf{A}_t = (a_t, a_{t+1}, \ldots, a_{t+H-1})$.
State-of-the-art VLAs~\citep{black2024pi_0, intelligence2025pi_} generate action chunks using conditional flow matching~\citep{lipman2022flow, zhan2026mean}: noise $\mathbf{A}^0_t \sim \mathcal{N}(0, I)$ is sampled and integrated through the learned velocity field $v_\pi$ from $\tau=0$ to $1$:
\begin{equation}
    \mathbf{A}_t^{\tau+\frac{1}{n}} = \mathbf{A}_t^\tau + \frac{1}{n} v_\pi( \mathbf{A}_t^\tau, \mathbf{o}_t, \tau),
    \label{eq:flow_step}
\end{equation}
where $\tau \in [0, 1)$ is the flow timestep and $n$ is the number of denoising steps.
During training, all actions within a chunk are corrupted at the same noise level $\tau$ and denoised together.

\textbf{Real-Time Chunking.}
Running modern VLAs with action generation time $\delta$ smaller than the controller period $\Delta t$ (typically 20$\sim$50\,ms) is challenging~\citep{black2025real}: even a 7B OpenVLA model optimized for speed~\citep{kim2025fine, kim2024openvla} requires $\delta = 321$\,ms on a server-grade A100 GPU.
The robot must therefore act while the policy is still computing, \ie, asynchronous execution.
Naively switching to a new action chunk as soon as it is ready, however, leads to inter-chunk discontinuity: adjacent chunks may correspond to different behavioral modes, producing jerky motions at chunk boundaries.

Real-time chunking (RTC)~\citep{black2025real} addresses this discontinuity by posing asynchronous execution as an \emph{inpainting} problem.
If inference takes $d$ (inference delay) steps to complete, the first $d$ actions of the new chunk are ``frozen'' to the values from the previous chunk before the new chunk becomes available.
The remaining actions must be generated to be consistent with this frozen prefix.
For flow-matching policies, RTC implements this inpainting via $\Pi$GDM~\citep{song2023pseudoinverse}, adding a gradient-based correction at each denoising step that steers the generation toward the target prefix $\mathbf{Y}$:
\begin{equation}
    \mathbf{g} = \left(\mathbf{Y} - \hat{f}_{\mathbf{A}^1}(\mathbf{A}^\tau)\right)^\top \text{diag}(\mathbf{W}) \cdot \frac{\partial \hat{f}_{\mathbf{A}^1}}{\partial \mathbf{A}'}\bigg|_{\mathbf{A}^\tau},
    \label{eq:guidance}
\end{equation}
where $\hat{f}_{\mathbf{A}^1}$ denotes the one-step denoising function and $\mathbf{W}$ is a vector of soft-masking weights.
The weights $\mathbf{W}$ assign $1$ to the first $d$ frozen actions, exponentially decay for the overlapping region, and $0$ for the final $s$ (execution horizon) actions.
During inference, the robot starts an inference every $s$ steps and executes the next $s$ actions from the most recent inference results.


\textbf{Discrete Diffusion Policies.} Discrete diffusion policies~\citep{liang2025discrete} tokenize continuous actions into discrete tokens via a vocabulary (\eg, VQ-VAE~\citep{VQVAE}, FAST~\citep{FAST} or $k$-bin quantization).
With a slight abuse of notation, we reuse $\mathbf{A}_t$ to denote the resulting action token sequence.
The forward process replaces tokens with a special $[\text{MASK}]$ token with a masking schedule,
while the reverse process iteratively executes unmasking operations conditioned on the current partially masked sequence.

A discrete diffusion policy is trained for one-step unmasking.
It consists of two components: $p_\pi$, which predicts the probability distribution over all token positions, and $f_\pi$, which samples which tokens to unmask at the current step from that distribution.
At each step $k$, the policy first predicts token identities via $p_\pi$ and then selects which tokens to unmask via $f_\pi$ given the observation $\mathbf{o}_t$:
\begin{equation}
    \mathbf{A}_t^{k+1} = \pi(\mathbf{A}_t^k, \mathbf{o}_t) = f_\pi\!\bigl(p_\pi(\mathbf{A}_t^k, \mathbf{o}_t)\bigr),
    \label{eq:dd_step}
\end{equation}
where $\mathbf{A}_t^k$ is the partially masked sequence at step $k$.
The discrete diffusion policy learns to predict the fully unmasked ground-truth action token sequence given the randomly masked token sequence.


\section{Flow-Matching is not Suitable for RTC}
\label{sec:limitations}


\begin{figure}
    \centering
    \includegraphics[width=\linewidth]{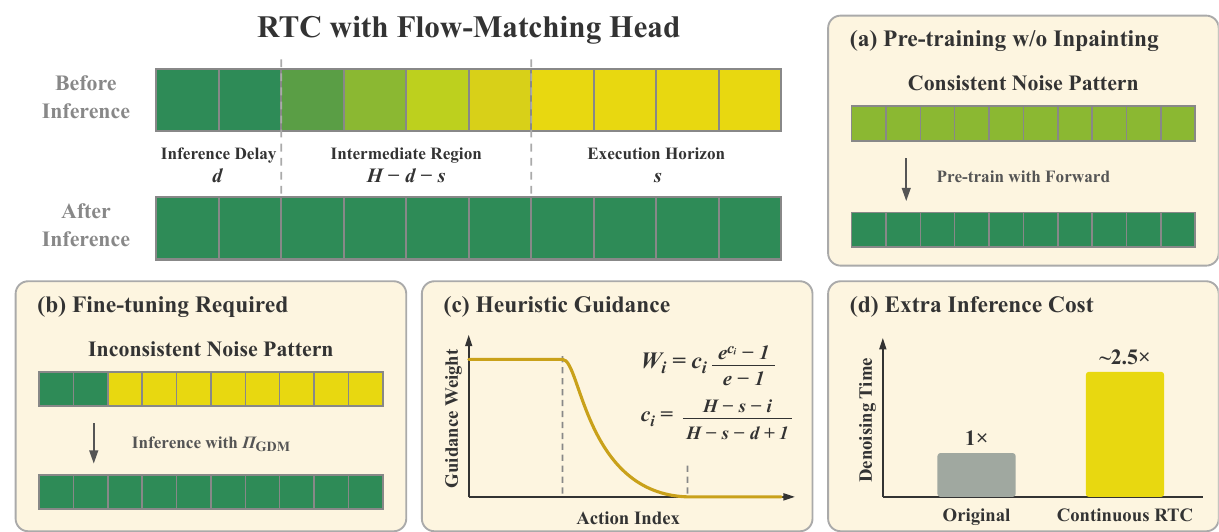}
    \caption{\textbf{RTC with flow-matching head.} Color represents the noise level, where green stands for the clear action and yellow stands for the pure noise. The flow-matching head is ill-suited for RTC because (a) during pre-training, the base policy is not trained on inpainting tasks; (b) to acquire this capability, a specially designed fine-tuning stage is required; (c) at inference time, RTC relies on heuristic guidance weight design; and (d) incurs extra cost for online correction computation.}
    \label{fig:continuousRTC}
    \vskip -0.15in
\end{figure}


\subsection{Pre-training without Inpainting}
\label{sec:lim_pretrain}
As illustrated in Figure~\ref{fig:continuousRTC}, pre-training of flow-matching policies~\citep{lipman2022flow} learns to generate conditioned on observations and a noised action chunk.
During training, all actions within a chunk are corrupted at the \textbf{consistent} noise level $\tau$ and are denoised together.
However, during RTC, the inpainting starts from an action chunk with \textbf{inconsistent} noise levels across positions: 0 for the committed actions, 1 for the new actions, and interpolated values for the intermediate actions.
Although the $\Pi$GDM~\citep{song2023pseudoinverse} guidance provides an external correction at inference time, the model has encountered neither such corrections nor the mixed-noise chunk pattern during the pre-training stage.
As a result, scaling up pre-training data or model capacity does not directly improve inpainting quality.

\subsection{Fine-tuning Required}
\label{sec:lim_finetuning}
Following Sec.~\ref{sec:lim_pretrain}, to improve the inpainting capability of the base policy, the mixed-noise chunk pattern must be explicitly introduced during training along with a corresponding inpainting objective.
This requires fine-tuning the model with an explicit inpainting loss~\citep{black2025training}, adding training complexity and risking interference with the base generation quality~\citep{wang2026real}.
This introduces a non-trivial additional engineering and computational burden, particularly for large VLAs that are expensive~\citep{hu2022lora} or even risky to fine-tune without degrading the base generation abilities~\citep{wang2026real}.

\subsection{Heuristic Guidance}
\label{sec:lim_guidance}

To effectively inpaint with flow-matching policies, online corrections like $\Pi$GDM~\citep{song2023pseudoinverse} are inevitable.
Completing Equation~\ref{eq:guidance}, the full denoising step of $\Pi$GDM is presented as follows:
\begin{equation}
\label{eqn: continuousRTC forward}
\mathbf{A}_t^{\tau+\frac{1}{n}} = \mathbf{A}_t^\tau + \frac{1}{n}\!\left(v_\pi(\mathbf{A}_t^\tau, \mathbf{o}_t, \tau) + \min\!\left(\beta,\, \frac{1-\tau}{\tau \cdot r_\tau^2}\right) \left(\mathbf{Y} - \hat{f}_{\mathbf{A}^1}(\mathbf{A}^\tau)\right)^\top \text{diag}(\mathbf{W}) \frac{\partial \hat{f}_{\mathbf{A}^1}}{\partial \mathbf{A}'}\bigg|_{\mathbf{A}^\tau} \right).
\end{equation}
Notably, the soft-masking weights $W$ in Equation~\ref{eqn: continuousRTC forward} follow a hand-crafted exponential decay schedule.
This important guidance schedule, which decides how to follow the previous action chunks, is fixed across different inference situations and is validated only through empirical studies.
This heuristic design not only creates a tuning burden, but also behaves inflexibly in response to different inference cases during asynchronous execution.

\subsection{Extra Inference Cost}
\label{sec:lim_cost}
Finally, running $\Pi$GDM at inference time requires non-negligible extra computation.
At each denoising step, computing the guidance term in Equation~\ref{eqn: continuousRTC forward} requires a vector-Jacobian product (VJP), which roughly doubles the per-step computation~\citep{wang2026real}. As a mechanism intended to enable smooth real-time execution, flow-matching-based RTC instead introduces additional inference delay, shrinking the time window available for generating new actions and counteracting its design goal.


\section{Discrete Diffusion Policies are Natural Asynchronous Executors}
\label{sec:method}

\begin{figure}
    \centering
    \includegraphics[width=\linewidth]{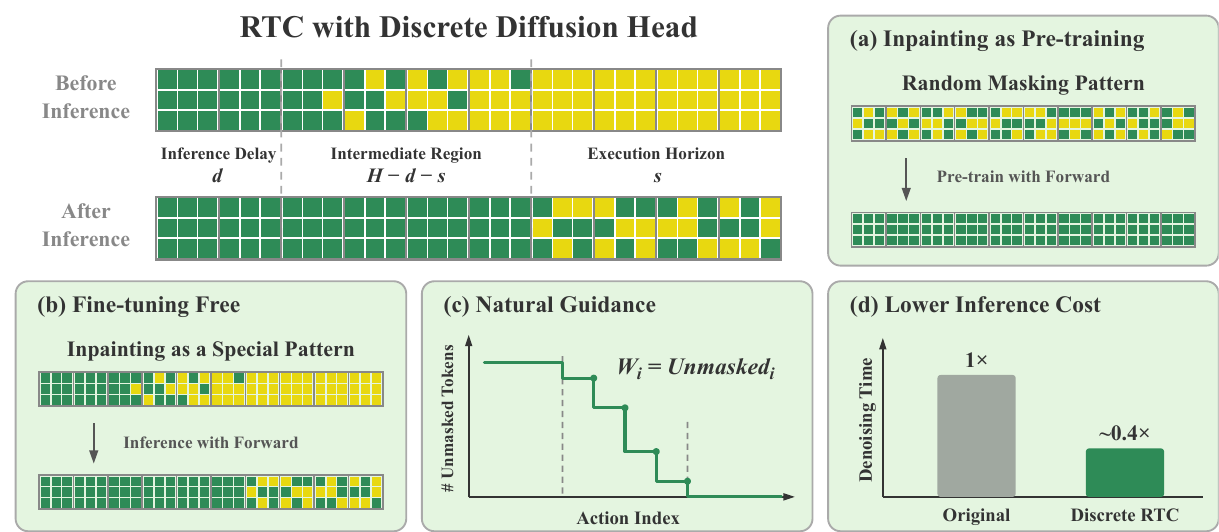}
    \caption{\textbf{RTC with discrete diffusion head.}  Color represents the masking status, where green stands for the unmasked token and yellow stands for the masked token. The discrete diffusion head is naturally suited for RTC because (a) during pre-training, the base policy is already trained on inpainting tasks; (b) consequently, no inpainting-specific fine-tuning is required; (c) at inference time, early stopping from the previous inference leaves a natural guidance signal; and (d) fewer tokens need to be unmasked per step, reducing the computational cost of each forward pass.}
    \label{fig:discreteRTC}
    \vskip -0.15in
\end{figure}

As illustrated in Figure~\ref{fig:discreteRTC}, we show that discrete diffusion policies can naturally resolve all the aforementioned limitations of flow-matching-based RTC at once.
\subsection{Learning to Inpaint from Pre-training without Fine-tuning}
\label{sec:method_pretrain}

In discrete diffusion, inpainting is the native operation: given a partially masked token sequence, the policy is asked to reconstruct the target action chunks by predicting the masked tokens.
Therefore, pre-training directly improves the model's ability to inpaint at inference time: scaling model, data, and pre-training computation yields better asynchronous execution without inpainting-specific fine-tuning.
To inpaint at inference time, no extra implementation, loss term, or inference-time guidance is needed:
\begin{equation}
    \label{eq:discrete_rtc_forward}
    \mathbf{A}_t^{k+1} = \pi(\mathbf{A}_t^k, \mathbf{o}_t).
\end{equation}
Unlike Equation~\ref{eqn: continuousRTC forward}, the model generates conditioned on the partially unmasked token sequence from the last inference, exactly as it was trained, by simply forwarding the base policy.

\subsection{Natural Guidance}
\label{sec:method_netural}

To solve the guidance issue, we exploit a key property of discrete diffusion: once a token is unmasked, it already carries a well-defined semantic meaning.
This stands in contrast to continuous diffusion or flow-matching, where the denoised action chunk becomes valid and executable all together only when the noise level reaches zero at the end of the denoising process.
Therefore, with discrete diffusion, the model can not only start inference from an intermediate state (e.g. a partially unmasked action chunk with prefix), but also terminate at an intermediate stage.

To ensure valid execution before the next inference, only the next $s$ actions after the committed actions are necessary to be fully unmasked. Therefore, the inference can early-exit when satisfying the minimal criteria, leaving the remaining actions a partially unmasked pattern. By carrying over the results, this pattern serves as a \emph{natural and adaptive} guidance for the next inference with the number of the unmasked tokens, replacing the heuristic and fixed guidance in flow-matching RTC.



\subsection{Lower Inference Cost}
\label{sec:method_lower}
Finally, since unmasking starts from an intermediate state and ends at an intermediate stage, keeping the number of tokens unmasked per step constant actually reduces the total number of tokens to unmask per inference to roughly $s / H$ of the original, or at least $1 - d/H$ by starting only from the actions to execute and terminating at complete action chunks.
This means that, instead of \emph{increasing} inference cost as in flow-matching-based RTC, our method can actually \emph{reduce} it by simply forwarding the base policy with fewer iterative steps.
Alternatively, we can keep the number of inference steps fixed and reduce the number of tokens unmasked per step, thereby producing finer-grained actions without increasing the action generation cost compared with generation from scratch.

\begin{figure}
    \centering
    \includegraphics[width=0.9\linewidth]{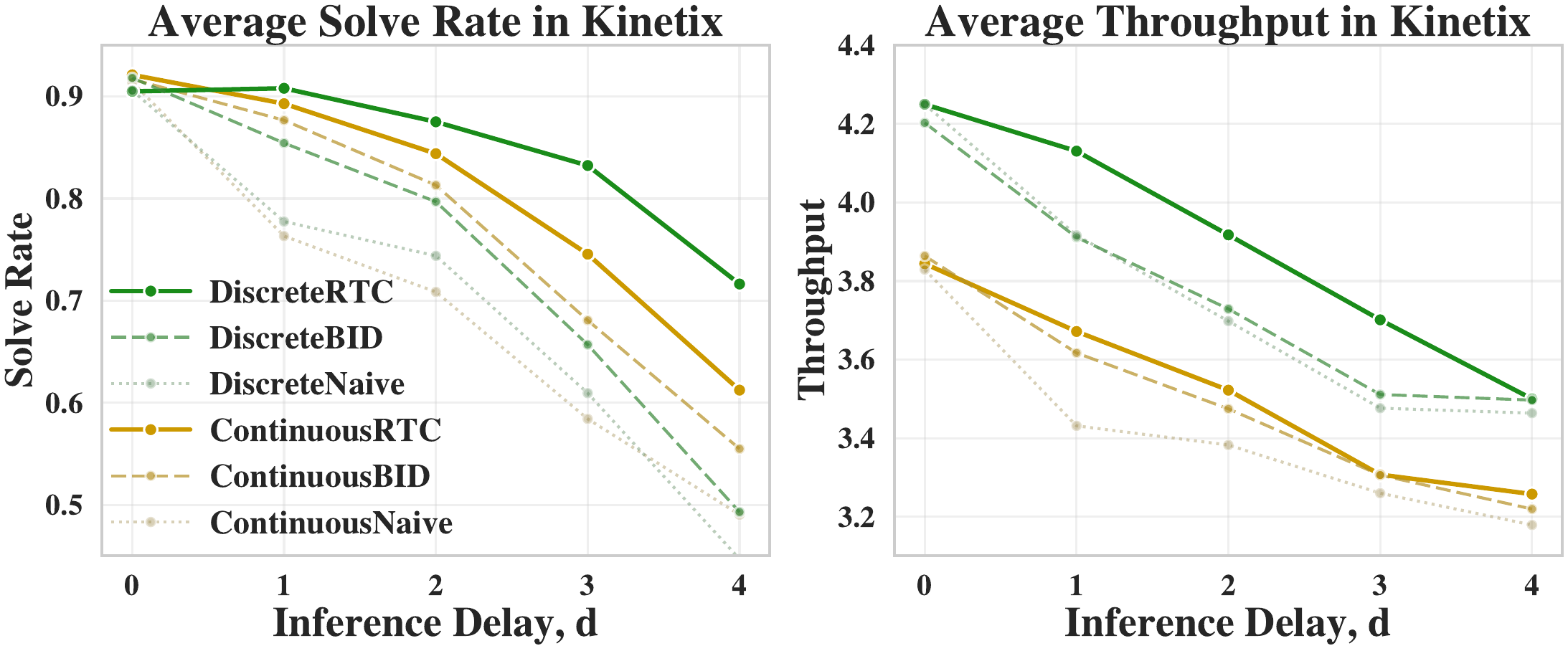}
    \caption{\textbf{Experimental Results in Kinetix.} The throughputs represent the task completed by the policy every 256 steps.
    \textbf{Left}: Average solve rate and throughputs across all environments with different inference delays;
    \textbf{Right}: Solve rates for every tasks with different inference delays.
    The execution horizon follows $s = \max(1, d)$ and each datapoint represents 2048 trials.}
    \label{fig:kinetix_main}
    \vskip -0.2in
\end{figure}

\section{Experiments}
\label{sec:experiments}



\subsection{Simulated Benchmark}
\label{sec:exp_sim}

We first evaluate DiscreteRTC in Kinetix~\citep{matthews2024kinetix}, a simulated benchmark of dynamic tasks.
To simulate imperfect actuation, we add Gaussian noise to the actions to make closed-loop corrections crucial for success~\citep{wang2026real}. Please refer to Appendix~\ref{app: addexp} for additional experimental results.

\textbf{Setup.}
We implemented the discrete diffusion policies upon the official codebase and datasets from RTC~\citep{black2025real}.
For fair comparison, we adopt the same training setup (optimizer, learning rate, epochs, etc.) and the same policy network size except the final layer for logits output.
We use a trivial $512$-bin quantization for the action chunk discretization.
Please refer to Appendix~\ref{app: simexp} for more details.

\textbf{Baselines.} With the same policy architecture, we compare against the flow-matching (Continuous) and the discrete diffusion (Discrete) action head under the following asynchronous strategies:
\begin{itemize}

    \item \textbf{Naive Async.} Chunks are generated from scratch and switched when the new one is ready.
    \item \textbf{Bidirectional decoding (BID)}~\citep{liu2024bidirectional}. Uses rejection sampling to maintain cross-chunk continuity. We use a batch size of $N = 16$ with no weak policy as the official implementation.
    \item \textbf{RTC}~\citep{black2025real}. Flow-matching policy with $\Pi$GDM inpainting and soft masking, representing the state-of-the-art asynchronous execution method for continuous policies.
\end{itemize}


\begin{figure}
    \centering
    \includegraphics[width=0.9\linewidth]{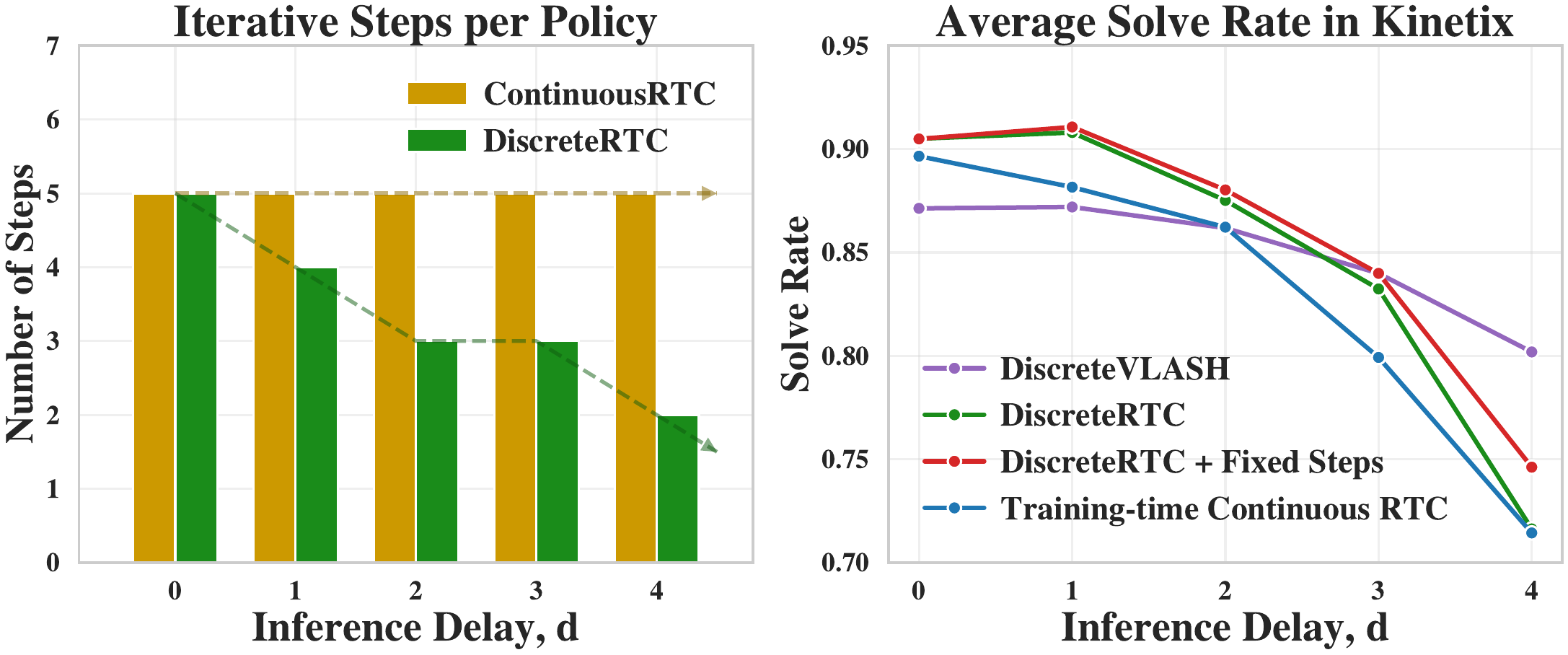}
    \caption{\textbf{Extended Experimental Results in Kinetix.}
    \textbf{Left}: Required iterative steps for each inpainting inference of different policy architectures in Kinetix with $s = \max(1, d)$;
    \textbf{Right}: Average solve rates of extended variants in Kinetix. The evaluation setup keeps the same with Figure~\ref{fig:kinetix_main}.
    }
    \label{fig:step_ablation}
    \vskip -0.2in
\end{figure}


\textbf{Results.}
We present the main results in Kinetix in Figure~\ref{fig:kinetix_main}.
Under different inference delays, DiscreteRTC consistently outperforms ContinuousRTC and other variants on both solve rates and throughputs, showing the advantage of the native inpainting capability in discrete diffusion policies.

We present the extended experimental results in Kinetix in Figure~\ref{fig:step_ablation}.
As shown in the results of iterative steps per policy, DiscreteRTC requires fewer iterative steps than ContinuousRTC, especially under large delays with execution horizon as discussed in Sec.~\ref{sec:method}.
Moreover, regarding the average solve rates of extended variants in Kinetix, we show three important observations:
(1) First, DiscreteRTC can outperform \textit{\textbf{Training-time ContinuousRTC}}~\citep{black2025training}, which is fine-tuned with an explicit objective and recipe to improve inpainting, further demonstrating the strong performance of the fine-tuning free DiscreteRTC.
(2) Next, by using fixed unmasking steps during inference, \textit{\textbf{DiscreteRTC + Fixed Steps}} can achieve better performance by generating more fine-grained actions with the same computation budget as discussed in Sec.~\ref{sec:method_lower}.
(3) Moreover, \textit{\textbf{DiscreteVLASH}}, which integrates the VLASH~\citep{tang2025vlash} can achieve more stable performance across varying delays but at the cost of performance drop under small delays.
This demonstrates that discrete diffusion can be seamlessly combined with advanced inference-time methods for further gains.

\subsection{Real-World Results}
\label{sec:exp_real}
Next, we validate DiscreteRTC on a real robot platform, focusing on dynamic tasks where inference latency and control frequency have a crucial impact on the closed-loop performance. Please refer to Appendix~\ref{app: realexp} for full deployment details and visualizations.

\textbf{Setup.}
We use a UR5e arm with a Robotiq gripper and a wrist-mounted RGB camera.
Developed with StarVLA~\citep{ye2026starvla}, all policies share a Qwen2.5-VL-3B-Instruct~\citep{yang2025qwen3} VLM paired with a layerwise cross-attention DiT~\citep{peebles2023scalable} action head.
The robot operates on a single RTX 4090.
Regarding the baselines, we mainly consider the Sync, RTC, and training-time RTC with flow-matching and discrete diffusion heads to better demonstrate our main claims.
We design three dynamic tasks to directly stress the reactiveness, \textbf{Dynamic Pick}, \textbf{Dynamic Place}, and \textbf{Hockey Defend}. Regarding the former two tasks, the robot is asked to pick a moving object or place the object on a moving platform; for the last, the robot is asked to defend a high-speed hockey attack from the opponent policy. These tasks require the robot to keep re-estimating the object pose and to execute agilely and smoothly.

\begin{figure}
    \centering
    \includegraphics[width=\linewidth]{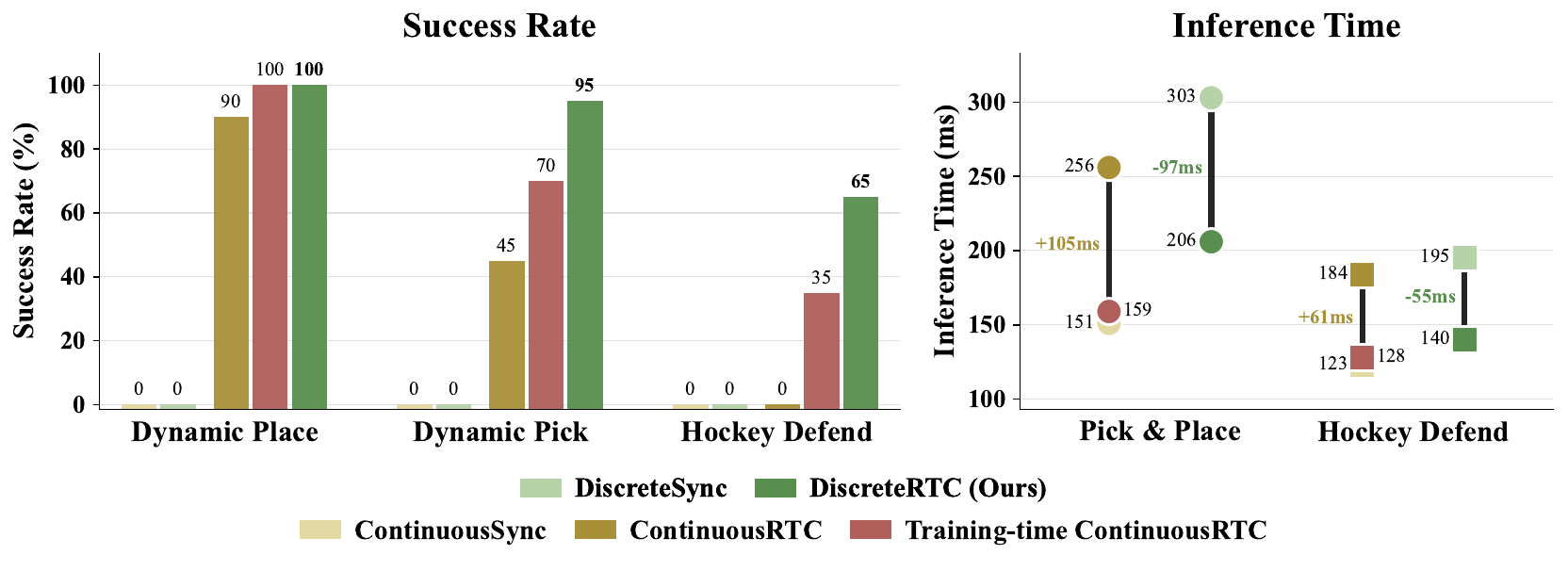}
    \caption{Real-world results on the three dynamic tasks. We report task success rate over 20 trials per task and average inference time (ms) on a single RTX 4090.
    }
    \label{fig:realworld_results}
    \vskip -0.2in
\end{figure}

\textbf{Results.}
Figure~\ref{fig:realworld_results} summarizes the real-world results.
Both Sync baselines completely fail ($0\%$), confirming that reactive asynchronous execution is indispensable in these regimes.
In terms of success rate, DiscreteRTC outperforms ContinuousRTC with a huge gap on the Hockey Defend and the Dynamic Pick where the reactiveness is critical for success.
In terms of inference time, the two asynchronous methods move in opposite directions: ContinuousRTC inflates flow-matching cost with a near-$1.7\times$ overhead from the $\Pi$GDM, whereas DiscreteRTC \emph{reduces} discrete diffusion cost to around $0.7\times$ since fewer tokens need to be unmasked during inference. Consequently, DiscreteRTC ends up both faster and better than ContinuousRTC.
Even compared with training-time ContinuousRTC with extra fine-tuning efforts, DiscreteRTC can still outperform with higher action qualities and success rates. However, the naive k-bin tokenization blocks DiscreteRTC from being more efficient than the training-time methods, a limitation discussed in the final section.



\section{Related Work}
\label{sec:related}

We briefly review the most relevant works in this section with an extended discussion in Appendix~\ref{app:related}.

\textbf{Efficient VLA via Discrete Diffusion.}
Prior efforts to accelerate VLAs span system-level optimization~\citep{ma2025running} and action tokenization~\citep{pertsch2025fast, liu2026oat, intelligence2025pi_}.
However, tokenization-based methods still rely on autoregressive decoding at inference time, limiting throughput.
Discrete diffusion~\citep{lou2023discrete} replaces autoregressive generation with parallel unmasking, and several recent works apply it to VLA action heads~\citep{liang2025discrete, wen2025llada, wen2025dvla, chen2025unified, ye2025dream, chen2026dfm, song2026fast}.
While these works demonstrate the efficiency of discrete diffusion, none investigate its structural advantage for asynchronous execution, which is the focus of ours.

\textbf{Asynchronous Execution Strategy.}
When inference latency exceeds the control period, the robot must act while computing~\citep{xiao2020thinking}.
With action chunking, strategies range from naive chunk switching and temporal ensembling~\citep{zhao2023learning} to rejection-based methods~\citep{liu2024bidirectional}.
Real-time chunking (RTC)~\citep{black2025real} poses chunk transitions as inpainting via $\Pi$GDM guidance~\citep{song2023pseudoinverse}, with follow-up works improving guidance design~\citep{yang2026abpolicy, lu2026faster} and learning inpainting at training time~\citep{black2025training, wang2026real, liu2026learning}.
However, all of these methods operate within the flow-matching framework,
while DiscreteRTC is the first to exploit the native inpainting capability of discrete diffusion policies for natural asynchronous execution.

\section{Discussion and Future Steps}
\label{sec:conclusion}
\textbf{Conclusions.}
We presented DiscreteRTC, which exploits the native inpainting capability of discrete diffusion for asynchronous real-time control, eliminating the need for inpainting-specific fine-tuning, heuristic guidance weights, and extra inference cost inherent to flow-matching-based RTC.
Simulated and real-world experiments show that DiscreteRTC outperforms ContinuousRTC and training-time baselines, and can be combined with methods such as VLASH for further gains.

\textbf{Limitations.}
We note several limitations of our current instantiation, each of which coincides with an active research direction that benefits DiscreteRTC.
(1)~Our naive $k$-bin quantization yields overly long token sequences, motivating compact, temporally aware tokenizers~\citep{pertsch2025fast, liu2026oat}.
(2)~Our modularized AR-VLM + discrete diffusion head prevents the backbone from participating in iterative unmasking, a gap that unified discrete-diffusion VLAs~\citep{wen2025llada, wen2025dvla, chen2025unified, ye2025dream} aim to close.
(3)~Traditional max-confidence unmasking does not yet translate the natural schedule into consistent gains.

\textbf{Future Steps.}
These limitations point to three concrete future research directions: 1) \emph{A time-causally ordered action tokenizer} that produces temporally ordered, compact token representations; 2) \emph{A unified discrete diffusion VLA} in which observation reasoning and action generation share the same backbone; and 3) \emph{An appropriate yet principled unmasking strategy} that fully unleashes the potential of the natural schedule. Emerging techniques such as AR-block decoding align better with the implicit autoregressive structure induced by RTC and integrate seamlessly.




\bibliography{example}  

\appendix
\section{Extended Related Works}
\label{app:related}

\textbf{Efficient VLA via Discrete Diffusion.}
To train and run VLAs efficiently, many prior efforts have focused on system-level optimization~\citep{kim2024openvla}, such as applying CUDA graph compilation and operator fusion to achieve real-time VLA inference on consumer GPUs~\citep{ma2025running}.
Action tokenization is another promising direction that leverages the fully optimized VLM pipeline by converting continuous actions into discrete tokens~\citep{intelligence2025pi_}.
Beyond trivial bin-based discretization, works like FAST~\citep{pertsch2025fast} and OAT~\citep{liu2026oat} compress action chunks into compact discrete token sequences, significantly reducing the required generation length.
However, these tokenization-based approaches still rely on autoregressive decoding at inference time, which inherently limits parallelism and throughput~\citep{intelligence2025pi_}; as a result, many of them fall back to a continuous flow-matching action head for deployment and are only used for efficient large-scale VLA pre-training.

Discrete diffusion~\citep{lou2023discrete} offers a promising alternative by replacing autoregressive generation with parallel unmasking.
Several recent works have begun exploring this direction:
\citet{liang2025discrete} apply discrete diffusion to the VLA action head, generating action chunks via iterative re-masking and parallel decoding;
LLaDA-VLA~\citep{wen2025llada} and DVLA~\citep{wen2025dvla} build unified discrete-diffusion VLA frameworks;
UD-VLA~\citep{chen2025unified} unifies observation reasoning and action generation within a single discrete diffusion backbone.
Among them, DreamVLA~\citep{ye2025dream} further replaces the VLM module with a discrete diffusion architecture, making it possible to pre-train the VLA with a unified backbone.

\textbf{Asynchronous Execution Strategy.}
When inference latency exceeds the control period, the robot must act while the policy is still computing.
The earliest formulation of this concurrent-control setting is Thinking While Moving~\citep{xiao2020thinking}, which executes the previous action during the current inference.
With action chunking~\citep{lai2025action}, the simplest strategy is naive asynchronous execution, which switches to a new chunk as soon as it is ready, but this can cause inter-chunk discontinuity.
Temporal ensembling~\citep{zhao2023learning} smooths consecutive chunks by averaging overlapping actions, at the cost of blurring distinct modes.
Bidirectional decoding (BID)~\citep{liu2024bidirectional} uses rejection sampling to maintain cross-chunk continuity, but at substantially higher computational cost.

Real-time chunking (RTC)~\citep{black2025real} elegantly poses chunk transitions as inpainting via $\Pi$GDM~\citep{song2023pseudoinverse} guidance, achieving strong performance on dynamic tasks.
Following RTC, several works have focused on improving inference-time inpainting through better guidance design:
ABPolicy~\citep{yang2026abpolicy} formulates the inpainting problem in a B-spline control-point action space for smooth manipulation.
Orthogonally, another line of follow-up works focuses on training-time methods that directly learn to inpaint during fine-tuning:
Training-Time RTC~\citep{black2025training} conditions on action prefixes during training to internalize the inpainting capability;
REMAC~\citep{wang2026real} follows a similar strategy but further eases the training--inference mismatch by mixing ground-truth and predicted actions during fine-tuning;
Legato~\citep{liu2026learning} learns a continuation-aware flow that reshapes the denoising dynamics to account for known prefixes at training time.
FASTER~\citep{lu2026faster} adopts a similar idea to rethink the indivisible characteristic of flow-matching training and inference, introducing a horizon-aware schedule to accelerate the time for the first available action.
There are also works that bridge both stages: VLASH~\citep{tang2025vlash} achieves lower policy-input temporal mismatch by estimating future robot states and pre-training the VLA to predict action chunks conditioned on both current observations and future states.

Our method, DiscreteRTC, is the first effort to step outside the flow-matching policy framework and analyze the asynchronous execution problem for discrete diffusion policies. It requires no training-time modification, yet achieves strong performance by exploiting the native inpainting capability of discrete diffusion---and can be combined with methods like VLASH~\citep{tang2025vlash} for further gains.


\newpage
\section{Additional Experiments}
\label{app: addexp}
\subsection{Natural Schedule Ablations}
\label{app: naturalmask}

\begin{figure}[h]
    \centering
    \includegraphics[width=\linewidth]{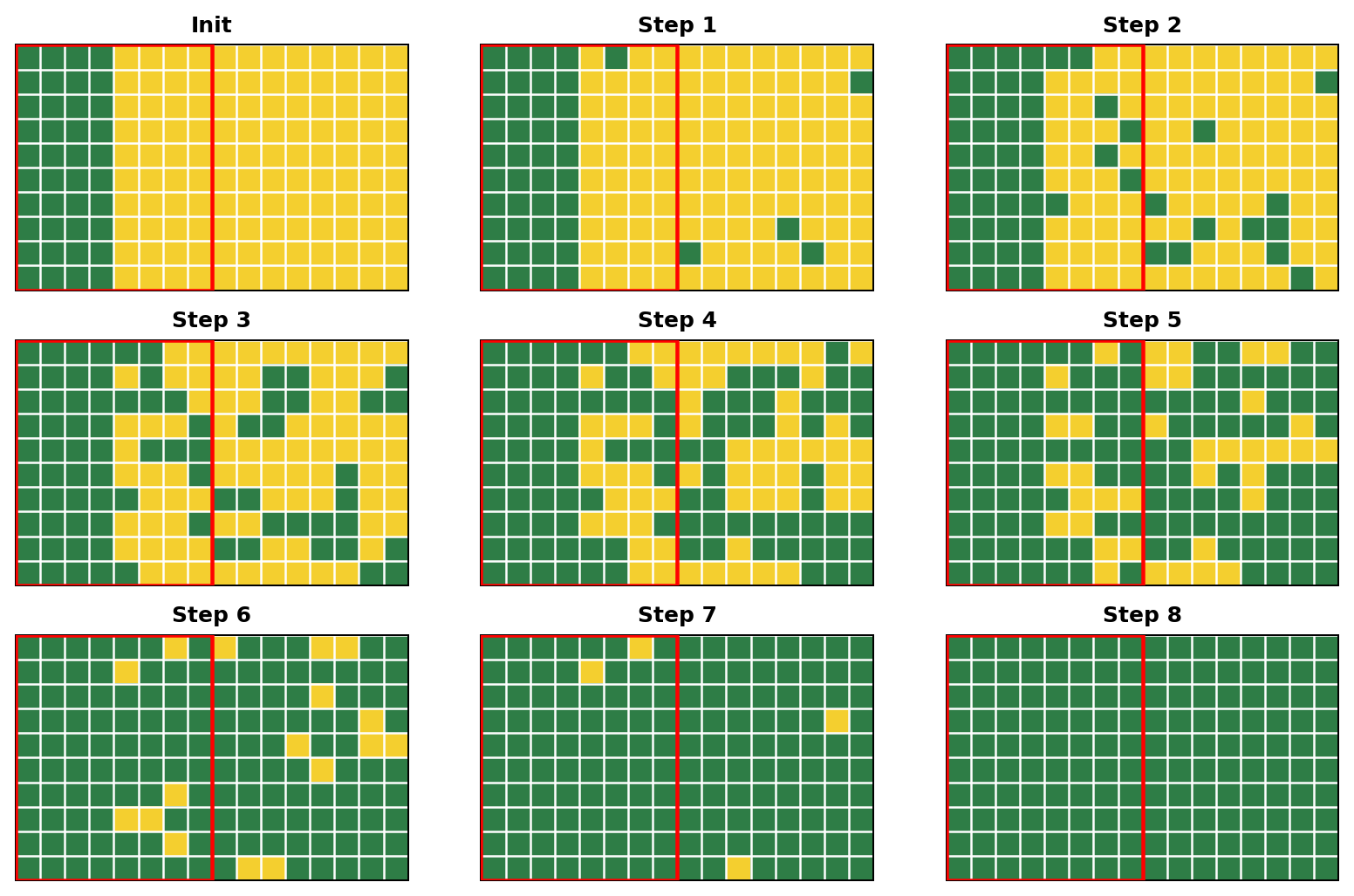}
    \caption{\textbf{Unmasking Trajectory Sample with Natural Schedule Inference.} Green blocks denote unmasked action tokens, yellow blocks denote masked tokens, and the red rectangle marks the early-stop boundary beyond which tokens do not need to be unmasked before the next inference.}
    \label{fig:natural_schedule}
\end{figure}

In practice, the natural schedule does work as expected compared to the simple hard mask approach. In this section, we show how the inappropriate unmasking-schedule limits the performance of natural schedule.
We visualize the per-step unmasking trajectory of the discrete diffusion policy under the max-confidence decoding strategy~\citep{liang2025discrete}.
In the visualization, green blocks represent unmasked action tokens, yellow blocks represent masked tokens, and the red rectangle marks the early-stop boundary: only tokens inside this boundary need to be fully unmasked before the next inference, while tokens outside can remain partially masked and carry forward as the natural guidance signal.
For each generation, the policy starts with the first $d=4$ actions already unmasked (the frozen prefix from the previous chunk) and is expected to inpaint the remaining tokens in an implicitly autoregressive fashion---unmasking the early actions first so that execution can begin as soon as possible.

However, as shown in Figure~\ref{fig:natural_schedule}, the max-confidence unmasking trajectory does \emph{not} exhibit this desired implicit AR-style ordering.
Instead, the policy distributes its unmasking budget across the entire chunk based on per-token confidence, and in most cases still consumes the full 8 unmasking steps before all tokens inside the early-stop boundary are resolved.
This both neutralizes the inference-cost savings promised by early stopping and degrades follow-up generations: because more tokens are fully unmasked by the time inference completes, the next chunk inherits a larger committed prefix and correspondingly less flexibility from the natural guidance signal.
This observation motivates the third direction in our future steps---designing a principled unmasking strategy (\eg, AR-block decoding) that aligns with the implicit autoregressive structure induced by RTC and fully unleashes the potential of the natural schedule.

\newpage

\subsection{Detailed Main Results in Kinetix}
\begin{figure}[h]
    \centering
    \includegraphics[width=\linewidth]{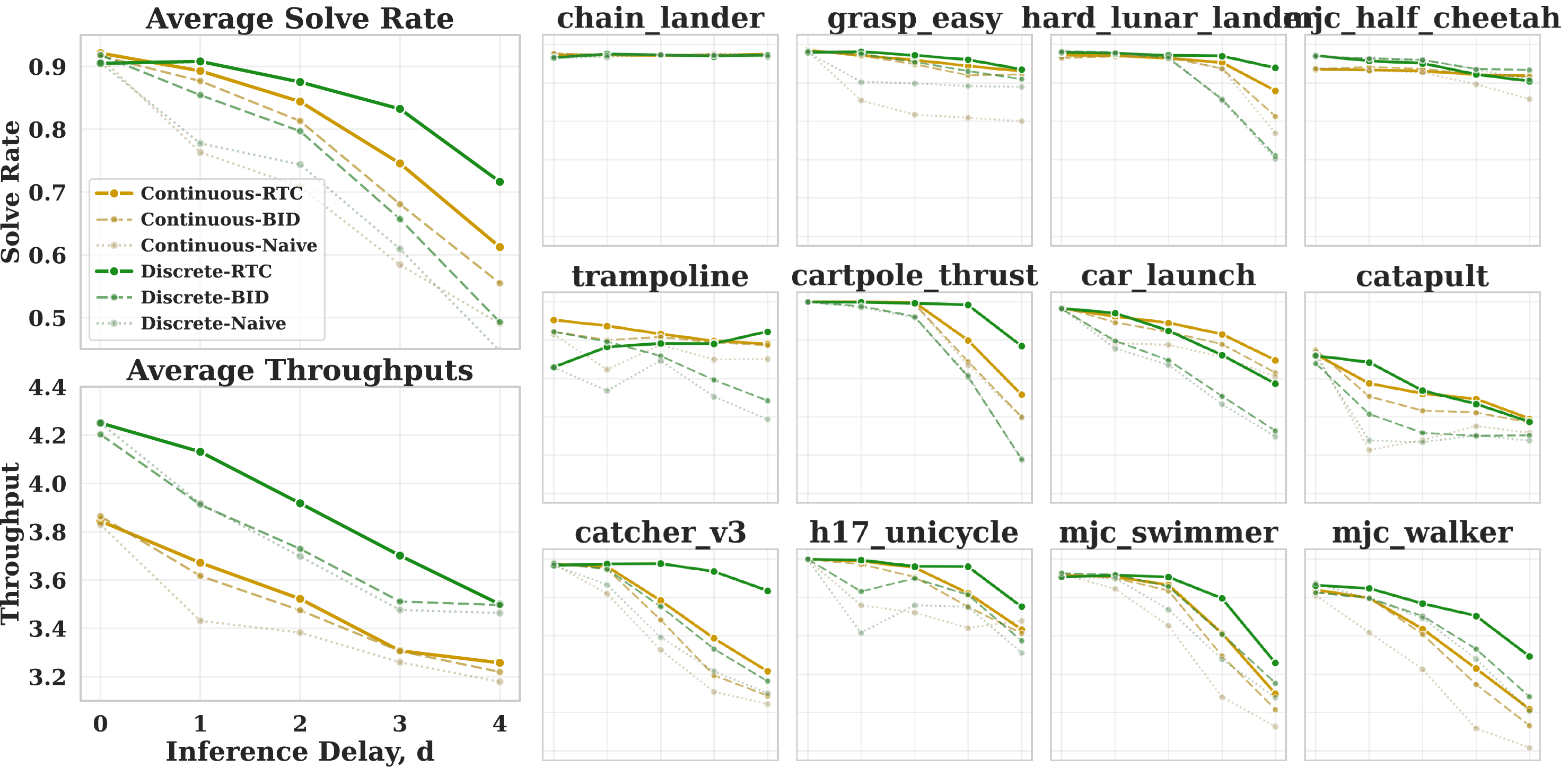}
    \caption{\textbf{Detailed Main Results in Kinetix.}  The evaluation setup keeps the same with Figure~\ref{fig:kinetix_main}. The small plots represent the results per Kinetix environment}
    \label{fig:ft_ablation1}
    \vskip -0.2in
\end{figure}

\subsection{Inpainting Fine-tuning}

\begin{figure}[h]
    \centering
    \includegraphics[width=\linewidth]{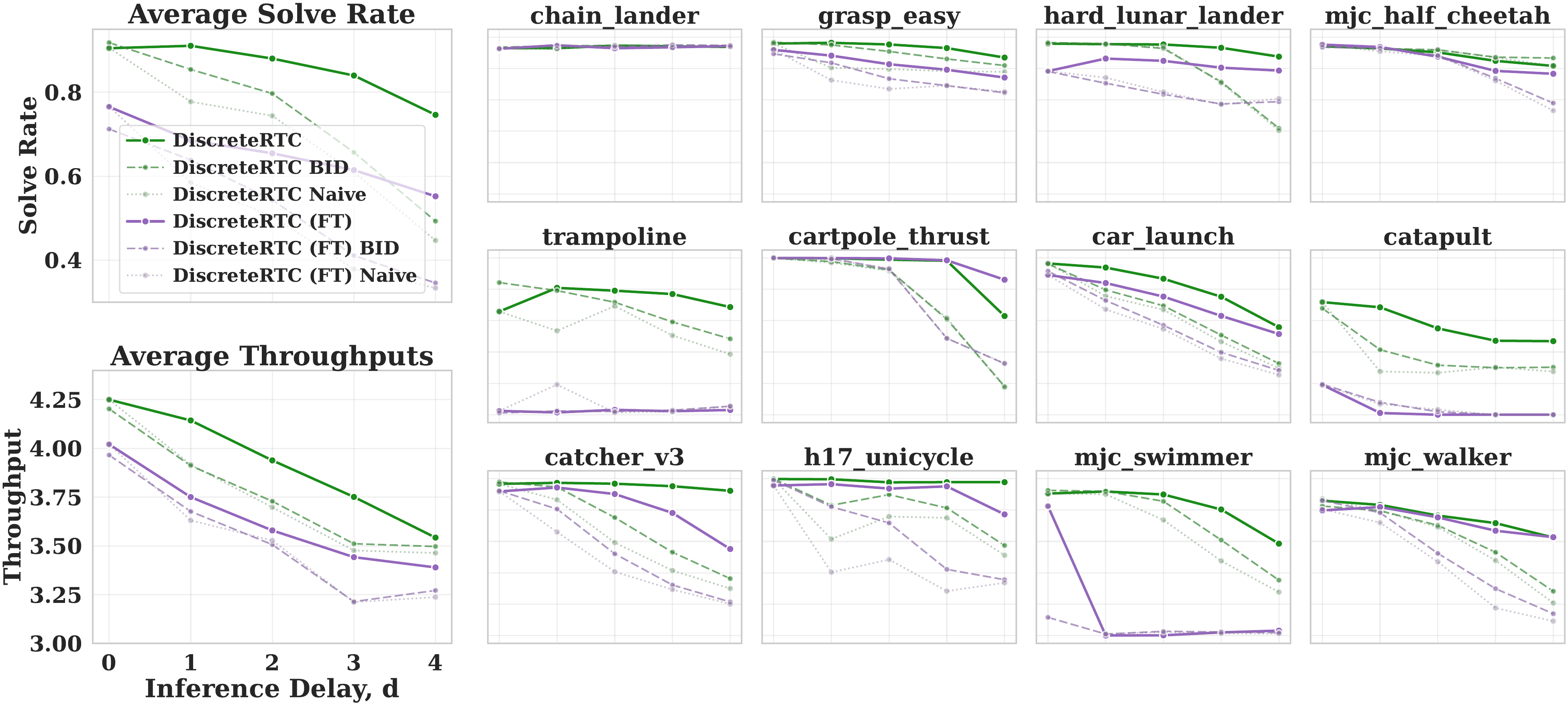}
    \caption{\textbf{Fine-tuning Ablation in Kinetix.} The evaluation setup keeps the same with Figure~\ref{fig:kinetix_main}. The small plots represent the results per Kinetix environment}
    \label{fig:ft_ablation2}
    \vskip -0.1in
\end{figure}

As an ablation of the fine-tuning-free claim, we explicitly fine-tune the pre-trained discrete diffusion policy with an inpainting-specific training recipe: for each training chunk we construct the target masking pattern by randomly masking as standard process first, then unmasking the prefix action tokens from $\{0, 1, 2, 3, 4\}$ to match the inference delay value distribution.
The intention is to bias the base policy toward the RTC-style prefix–continuation and further improve inpainting quality.

However, as shown in Figure~\ref{fig:ft_ablation1} and ~\ref{fig:ft_ablation2}, this naive fine-tuning strategy brings a consistent performance drop across all 12 Kinetix levels and every inference delay. This suggests that a naive prefix-unmasking recipe can actually \emph{hurt} the native inpainting capability of the pre-trained discrete diffusion policy. Improving beyond
the fine-tuning-free DiscreteRTC therefore appears to require either a carefully designed inpainting-aware fine-tuning recipe that does not distort the pre-training distribution, or just directly scaling the base policy's pre-training.

\newpage
\section{Kinetix Implementation Details}
\label{app: simexp}
The codebase is developed upon the \href{https://github.com/Physical-Intelligence/real-time-chunking-kinetix}{official RTC codebase}~\citep{black2025real}, implemented in JAX~\citep{jax2018github} with Flax NNx on 8 RTX 6000 PRO.

\textbf{Environment.}
We use Kinetix~\citep{matthews2024kinetix} (\texttt{Kinetix-Symbolic-Continuous-v1}, \texttt{large} configuration).
The physics simulation runs at 60\,Hz with frame skip 2, yielding 30\,Hz control.
Actions are normalized to $[-1, 1]$ with Gaussian noise (std 0.1) added during both data collection and evaluation.
Observations include a 4-frame history stack and the previous action.

\textbf{Policy Architecture.}
Both policies share an MLPMixer backbone with AdaLN conditioning: 4 blocks, each with token-mixing (hidden dim 64) and channel-mixing (hidden dim 512) paths, operating on $H{=}8$ tokens (one per chunk timestep) with channel dimension 256.

For \textbf{flow-matching policy}, the input is the observation concatenated with the noised action chunk.
A sinusoidal time embedding conditions the AdaLN layers, and the output is a velocity vector.
For \textbf{discrete diffusion policy}, actions are discretized into 512 bins via uniform quantization.
The input is the observation concatenated with bin embeddings of the partially masked token sequence.
The token sequence is packed from $(H, \text{action\_dim}, \text{embedding\_dim})$ to $(H, \text{channel\_dim})$ via a linear layer so the mixer sees the same 8-token structure.
A learned vector replaces the time conditioning.

\textbf{Training.}
Both policies are trained for 32 epochs with batch size 512 using AdamW.
The flow-matching policy uses a constant LR schedule and MSE loss on the predicted velocity field.
The discrete diffusion policy uses cosine LR decay and cross-entropy on masked positions plus an auxiliary L1 loss ($\lambda{=}0.1$) with a cosine masking schedule.
The training-time RTC is trained with the same setting with its simulated delay randomly sampled from the possible set, following the RTC codebase. The starting checkpoint is selected from epoch 24, and finetuned for 8 more epochs.

\textbf{Evaluation.}
Both policies use 5 denoising/unmasking steps.
For ContinuousRTC, $\Pi$GDM guidance with exponential-decay soft masking and clipping threshold $\beta$ steers generation toward the frozen prefix.
For DiscreteRTC, the frozen prefix tokens are directly placed, and for parallel evaluation across the environments, we use a hard mask for the chunk continuation.
All remaining hyperparameters are listed in Table~\ref{tab:hyperparams_sim}.

\begin{table}[h]
\centering
\caption{Simulation experiment hyperparameters.}
\label{tab:hyperparams_sim}
\begin{tabular}{lcc}
\toprule
\textbf{Hyperparameter} & \textbf{Flow-Matching} & \textbf{Discrete Diffusion} \\
\midrule
\multicolumn{3}{l}{\textit{Architecture}} \\
Channel / channel hidden / token hidden dim & \multicolumn{2}{c}{256 / 512 / 64} \\
Number of MLPMixer layers & \multicolumn{2}{c}{4} \\
Action chunk size ($H$) & \multicolumn{2}{c}{8} \\
Number of bins / embedding dim & --- & 512 / 128 \\
\midrule
\multicolumn{3}{l}{\textit{Training}} \\
Batch size / epochs & \multicolumn{2}{c}{512 / 32} \\
Optimizer / LR / weight decay & \multicolumn{2}{c}{AdamW / $3 \times 10^{-4}$ / $10^{-2}$} \\
Gradient norm clip & \multicolumn{2}{c}{10.0} \\
LR schedule (warmup steps) & Constant (1{,}000) & Cosine decay (2{,}000) \\
Training loss & MSE (velocity) & CE + L1 ($\lambda{=}0.1$) \\
Training mask schedule & --- & Cosine \\
\midrule
\multicolumn{3}{l}{\textit{Evaluation}} \\
Denoising / unmasking steps & 5 & 5 \\
Decode temperature / choice temperature & --- & 1.0 / 0.0 \\
RTC guidance & $\Pi$GDM ($\beta{=}5.0$) & Hard Mask \\
Inference delays ($d$) & \multicolumn{2}{c}{\{0, 1, 2, 3, 4\}} \\
Execution horizon ($s$) / trials & \multicolumn{2}{c}{$\max(1, d)$ / 2{,}048} \\
\bottomrule
\end{tabular}
\end{table}

\newpage
\section{Real-world Implementation Details}
\label{app: realexp}

\begin{figure}[h]
    \centering
    \includegraphics[width=0.9\linewidth]{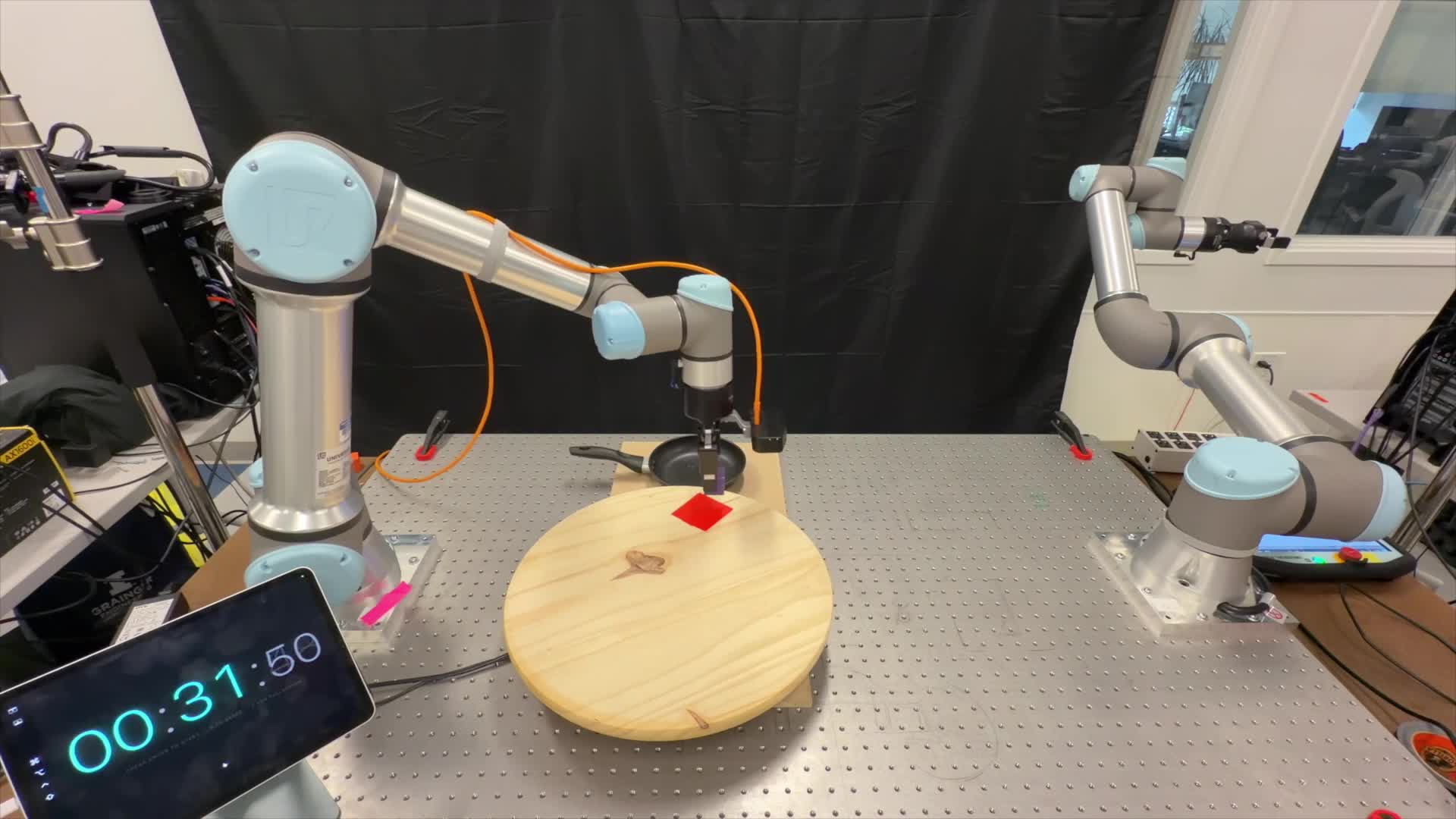}
    \caption{Dynamic Pick and Place Real-world Setup}
    \label{fig:real_setup}
\end{figure}

\begin{figure}[h]
    \centering
    \includegraphics[width=0.9\linewidth]{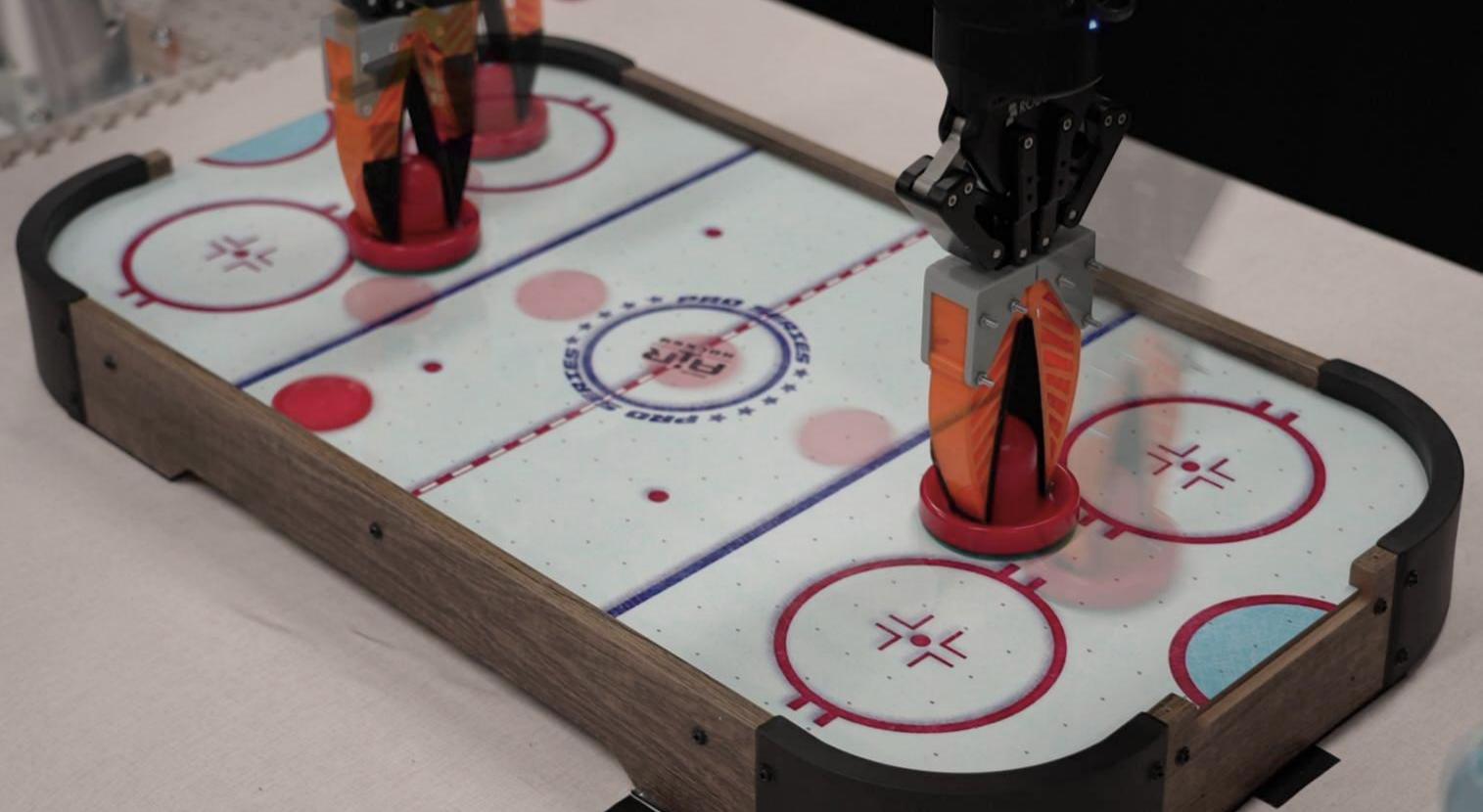}
    \caption{Hockey Defend Real-world Setup}
    \label{fig:hockey_setup}
\end{figure}

\textbf{Hardware and Data.}
We use a single UR5e arm with a Robotiq gripper and a wrist-mounted RGB camera.
Demonstrations are recorded at 500\,Hz via the FastUMI pipeline.
Each action is a 10D vector $[\Delta x, \Delta y, \Delta z, \text{rot6d}(6), \text{gripper}]$, with translational dimensions normalized to $[-1, 1]$ via min-max scaling, rotation dimensions left unnormalized, and the gripper binarized to $\{0, 1\}$.
Images are center-cropped and resized to $224 \times 224$.
Training uses 8 RTX Pro 6000 with DeepSpeed ZeRO-2 and gradient checkpointing; deployment runs on a single RTX 4090.

\textbf{Policy Architecture.}
All policies use Qwen2.5-VL-3B-Instruct~\citep{yang2025qwen3} as the VLM backbone.
Both action heads share the same QwenPI-style layerwise cross-attention DiT~\citep{peebles2023scalable}: 36 blocks, each with AdaLN, cross-attention to the corresponding VLM hidden states, self-attention, and a feedforward network.
Both prepend 32 learnable future vision tokens and add positional embeddings.
When proprioceptive state is available, a two-layer MLP encodes it and prepends it to the sequence.

For \textbf{flow-matching head}, a three-layer MLP fuses per-timestep action embeddings with sinusoidal flow-time embeddings (output dim 2048); a sinusoidal \texttt{TimestepEncoder} provides AdaLN conditioning; and a two-layer MLP decodes the velocity output.
For \textbf{discrete diffusion head}, actions are quantized into 256 bins per dimension, yielding $16 \times 10 = 160$ tokens per chunk.
Tokens (including a learnable \texttt{[MASK]}) are embedded via \texttt{nn.Embedding}(257, 2048) with no timestep conditioning.
For the training-time RTC, we select the final checkpoint to be continuously trained for another 2500 steps with the same setting. The simulated delay is pre-fixed with the estimated delays, $d=4$ for the dynamic pick \& place and $d=6$ for the hockey defend. The RTC parameters here and below are selected to match the exact inference delay in deployment instead of manually adjusting.


\textbf{Dynamic Pick \& Place Deployment.}
The robot operates in closed-loop at 20\,Hz with 5$\times$ linear interpolation producing a 100\,Hz servo stream.
Camera capture and model inference run in a background thread, overlapping with action execution.
The rotation speed of the turntable for dynamic pick and place are 15 and 20 rpm.
The discrete diffusion head uses 8 MaskGIT unmasking steps with a cosine decode schedule (decode temperature 0.0, choice temperature 0.1).
By default we use non-fixed unmasking steps for the discrete diffusion policies and running RTC with $d=4$, $s=4$ for discrete diffusion policies, $d=5$, $s=5$ for flow-matching policies.

\textbf{Air Hockey Defend Deployment.}
The robot operates in closed-loop at 50\,Hz with 2$\times$ linear interpolation producing a 100\,Hz servo stream to match the higher inference speed requirements. The hardware setup shares the same with the Dynamic Pick \& Place.
By default we use non-fixed unmasking steps for the discrete diffusion policies and running RTC with $d=7$, $s=7$ for discrete diffusion policies, $d=9$, $s=9$ for flow-matching policies.

All hyperparameters are summarized in Table~\ref{tab:real_hyperparams}.

\begin{table}[h]
\centering
\caption{Real-world experiment hyperparameters.}
\label{tab:real_hyperparams}
\begin{tabular}{lcc}
\toprule
\textbf{Hyperparameter} & \textbf{Flow-Matching} & \textbf{Discrete Diffusion} \\
\midrule
\multicolumn{3}{l}{\textit{VLM Backbone \& Action Head}} \\
VLM backbone & \multicolumn{2}{c}{Qwen2.5-VL-3B-Instruct} \\
Action head & \multicolumn{2}{c}{Layerwise Cross-Attention DiT} \\
DiT layers / heads / head dim & \multicolumn{2}{c}{36 / 32 / 64 (inner dim 2048)} \\
Dropout / future vision tokens & \multicolumn{2}{c}{0.2 / 32} \\
Action encoder & 3-layer MLP + sinusoidal time & Embedding(257, 2048) \\
Action decoder & 2-layer MLP (hidden 1024) & Linear $\rightarrow$ 256 logits \\
Timestep conditioning & TimestepEncoder (AdaLN) & None (zero vector) \\
\midrule
\multicolumn{3}{l}{\textit{Action Representation}} \\
Action dim &  \multicolumn{2}{c}{10 (Pick \& Place) / 2 (Hockey)} \\
chunk length $H$ &  \multicolumn{2}{c}{16 (Pick \& Place) / 32 (Hockey)}\\
Representation & Continuous & 256-bin quantization \\
\midrule
\multicolumn{3}{l}{\textit{Training}} \\
Optimizer & \multicolumn{2}{c}{AdamW ($\beta_1{=}0.9, \beta_2{=}0.95$)} \\
LR (VLM / action model) & \multicolumn{2}{c}{$10^{-5}$ / $10^{-4}$} \\
LR scheduler (warmup / total steps) & \multicolumn{2}{c}{Cosine, min $5{\times}10^{-7}$ (5k / 30k)} \\
Batch size (per device $\times$ GPUs) & \multicolumn{2}{c}{$8 \times 8 = 64$} \\
Mixed precision & \multicolumn{2}{c}{bfloat16 (VLM) / float32 (action head)} \\
Training loss & MSE (velocity) & CE + L1 ($\lambda{=}0.1$) \\
Noise / mask schedule & Beta($1.5, 1.0$), $s{=}0.999$ & Cosine \\
\midrule
\multicolumn{3}{l}{\textit{Inference \& Deployment}} \\
Denoising / unmasking steps & 8 & 8 \\
Decode / choice temperature & --- & 0.0 / 0.1 \\
Control / servo frequency & \multicolumn{2}{c}{20\,Hz / 100\,Hz ($5\times$ interpolation)} \\
Robot / camera & \multicolumn{2}{c}{UR5e + Robotiq / wrist-mounted fish-eye RGB} \\
\bottomrule
\end{tabular}
\end{table}

\end{document}